\documentclass{article}
\usepackage[nonatbib,preprint]{neurips_2019}
 \usepackage{neurips_2019}

\usepackage[utf8]{inputenc} 
\usepackage[T1]{fontenc}    
\usepackage{hyperref}       
\usepackage{url}            
\usepackage{booktabs}       
\usepackage{amsfonts}       
\usepackage{nicefrac}       
\usepackage{microtype}      

\usepackage{amsmath}
\usepackage{graphicx}
\usepackage{xcolor}

\title{Hierarchical Associative Memory}

\author{
  Dmitry Krotov \\
  MIT-IBM Watson AI Lab \\
  IBM Research\\
  \texttt{krotov@ibm.com} \\
}

\begin{document}
\maketitle

\begin{abstract}
  Dense Associative Memories or Modern Hopfield Networks have many appealing properties of associative memory. They can do pattern completion, store a large number of memories, and can be described using a recurrent neural network with a degree of biological plausibility and rich feedback between the neurons. At the same time, up until now all the models of this class have had only one hidden layer, and have only been formulated with densely connected network architectures, two aspects that hinder their machine learning applications. This paper tackles this gap and describes a fully recurrent model of associative memory with an arbitrary large number of layers, some of which can be locally connected (convolutional), and a corresponding energy function that decreases on the dynamical trajectory of the neurons' activations. The memories of the full network are dynamically ``assembled'' using primitives encoded in the synaptic weights of the lower layers, with the ``assembling rules'' encoded in the synaptic weights of the higher layers. In addition to the bottom-up propagation of information, typical of commonly used feedforward neural networks, the  model described has rich top-down feedback from higher layers that help the lower-layer neurons to decide on their response to the input stimuli.
  \end{abstract}

\section{Introduction}
Discrete \cite{Hopfield_82} and continuous \cite{Hopfield_84} Hopfield Networks are recurrent neural networks with dynamical trajectories converging to fixed point attractor states and described by an energy function. While having many desirable properties of associative memory, these classical systems suffer from a small memory storage capacity, which scales linearly with the number of input features \cite{Hopfield_82}. 

Dense Associative Memories or Modern Hopfield Networks   \cite{Krotov_Hopfield_2016, Demircigil, Hochreiter, Krotov_Hopfield_2020} are generalizations of classical Hopfield Networks that break the linear scaling relationship between the number of input features and the number of stored memories. This is achieved by introducing stronger non-linearities (neurons' activation functions)  leading to super-linear (even an exponential) memory storage capacity as a function of the number of feature neurons (the network still requires a sufficient number of hidden neurons \cite{Krotov_Hopfield_2020}). This property makes Modern Hopfield Networks appealing tools for many problems in machine learning and cognitive and neuro-sciences. At the same time, up until now all the models of this class have had only one hidden layer and dense connectivity, and, for this reason, have not had the representational richness and inductive biases required by most machine learning problems.   

This paper tackles this gap and describes a generalization of the Modern Hopfield Networks to an {\it arbitrary large number of hidden layers}, and networks with local connectivity. First, using the method of Lagrangian functions \cite{Krotov_Hopfield_2020}, I derive a general formulation of the Modern Hopfield Network with full all-to-all connectivity, so that every neuron can have its own activation function and its own kinetic time constant defining the dynamics. This network is described by a system of coupled non-linear differential equations and an energy function that decreases on the solution to those equations. Second, the network has been given a layered structure so that neurons within each layer have the same activation function and the same time constant. I derive explicit dynamical equations, their corresponding energy functions, and formulate the convergence property of these hierarchical layered networks in terms of their Lagrangians and energy functions. Unlike previously published methods, the formulation with the Lagrangian functions makes it possible to build multilayer Hopfield Networks with arbitrary activation functions that can depend on the activities of groups of neurons (as opposed to neuron-wise non-linearities). This is particularly important considering the recent correspondence \cite{Hochreiter} relating Dense Associative Memories to the attention mechanism in Transformers, which requires contrastive normalization in the hidden layer \cite{Krotov_Hopfield_2020} with the softmax activation. Another advantage of the method described in this work is that it does not require the inversion of the activation functions, which is typically used for deriving the energy (Lyapunov) functions, see for example \cite{Hopfield_84, Xie_Seung}. Such inverse activation functions are often undefined in the architectures used in practice, e.g. models with softmax attention.     

In the conventional feedforward deep neural networks (e.g. multi-layer perceptrons, CNNs, Transformers, etc.), the information flows unidirectionally from the input layer through a sequence of hidden layers to the output layer. In contrast, the Hierarchical Associative Memory (HAM) model considered in this work is a {\it fully recurrent neural network} (see the connectivity diagram in Fig.\ref{Fig:architecture}), which enables both {\it bottom-up flow of information} (inputs activate hidden neurons in higher layers), and {\it top-down flow of information} (activation patterns of the hidden neurons in higher layers dictate the temporal evolution of the neurons' states in the lower layers). The existence of the global (describing the entire network as opposed to individual layers) energy function, which is bounded from below, restricts the many possible types of trajectories permitted by such non-linear recurrent networks (e.g. fixed points, limit cycles, chaotic behavior, etc.) to fixed points only. Importantly, the existence of the global energy function requires that the matrix of feedback weights is equal to the transposed matrix of the feedforward weights. 

In the Modern Hopfield Networks with one hidden layer \cite{Krotov_Hopfield_2016, Demircigil, Hochreiter, Krotov_Hopfield_2020} the fixed points of the dynamical trajectories (memories) correspond to synaptic weights of individual hidden units (or averages across a subset of hidden units). In contrast, in the HAM model the attractor states are constructed representations that are ``assembled'' from primitives encoded in the synaptic weights of lower layers, with ``assembling rules'' encoded in the synaptic weights of the higher layers. Importantly, the primitives of the lower layers can be reused in multiple memories. This modular organization of the attractors makes HAMs appealing tools in situations requiring modelling hierarchical data (e.g. images or graphs), when existence of the attractors in the space of neuron's configurations is desirable.  

The connectivity diagram and the energy function of the proposed HAM model may look reminiscent of the Deep Restricted Boltzmann Machines (DRBMs) \cite{DRBM} and Deep Belief Networks (DBN) \cite{DBN}. There are several conceptual differences between them. First, both DRBMs and DBNs are probabilistic systems with stochastic evolution of states given by Markov chains sampling. In contrast, HAMs are attractor networks with completely deterministic temporal dynamics. Second, DBNs are typically trained using layer-wise learning methods (e.g. contrastive divergence) \cite{CD}, while HAMs can be trained end-to-end using for example the backpropagation through time algorithm. Additionally, the formulation with the Lagrangian functions used in this work makes it possible to define the energy function for a broader class of activation functions than those that are typically considered in the RBM literature. 

A layered arrangement of neurons in the continuous state Hopfield network with symmetric weights have been studied by Xie and Seung \cite{Xie_Seung}, which is the closest previously published approach to the HAM model discussed in this work. The dynamical equations and the energy function in \cite{Xie_Seung} were derived on the assumption that the activation function for each individual neuron is a non-linear function of the activity of that neuron only (no contrastive or divisive normalizations are allowed). Note that most of the Modern Hopfield Networks studied so far do not satisfy these assumptions. For instance, using the terminology of \cite{Krotov_Hopfield_2020} model B uses contrastive normalization in the hidden layer, i.e. requires the softmax activation function that depends on all the neurons in that layer. Model C uses divisive normalization in the feature layer. Thus, the approach of \cite{Xie_Seung} cannot be used to describe these models. Additionally, the energy function of \cite{Xie_Seung} requires the inversion of the activation function. Note that all these limiting cases can be easily handled by the formalism with the Lagrangian functions used in this work, and the model of \cite{Xie_Seung} can be derived as a limiting case of the HAM model.  

Another line of work describes the energy function of associative memory as a neural network, and performs pattern retrieval using a gradient descent on that energy with respect to the inputs \cite{Bartunov}. The ``read'' step in that approach involves a non-local (involving many-body interactions between all the neurons) gradient-descent-based computation for retrieving the patterns from the memory. For this reason, the retrieval step cannot be interpreted as a computation performed by a local neural network.  In contrast, the HAM approach begins with a local recurrent neural network that performs the retrieval step. The energy function arises as a mathematical consequence of this architectural choice. In addition to being fully analytically tractable, the HAM approach has the advantage of being in the same universality class as Modern (or classical) Hopfield Networks when it comes to its biological plausibility - both Hopfield Networks and HAM use local computation for memory retrievals. 

Another related idea is the neural ordinary differential equations (ODE) \cite{NeurODE}, which describes the state of a neural network in continuous time using a system of non-linear differential equations. The difference with the HAM model is that neural ODEs typically do not have an energy function that describes their solution. The same applies to Deep Equilibrium Models \cite{DEQ}, they converge to a fixed point state, but do not necessarily have an underlying energy function.   

From the biological perspective most cortical areas of the brain have a rich set of feedback connections \cite{Felleman}. The computational role of these connections is  poorly understood. The existence of the energy function in the HAM model requires that the feedback connections are described by the transposed matrix of the feedforward connections. Although there is no known reason for this constraint to be satisfied in real biological networks, the introduction of feedback connections places the HAM model one step closer to the biological reality compared to models of the cortex based on feedforward architectures, like CNNs or Transformers. 

Last but not least, there is a rich literature studying the relationship between the Modern Hopfield Networks (Dense Associative Memories) and RBMs from the perspective of statistical physics. For instance, \cite{Agliari} studies the effect of synaptic noise on the memory storage and retrieval. A class of non-quadratic energy functions related to Dense Associative Memories called the Relativistic Hopfield Model was studied in \cite{rel_HM, Barra}. See also a recent review \cite{Marullo} for a more systematic discussion of these results.

\section{A General Formulation of the Modern Hopfield Network}\label{FC:Section} 
Biological neural networks have a large degree of heterogeneity in terms of different cell types \cite{RamonyCajal, Heterogeneity}. This section describes a mathematical model of a fully connected associative memory network assuming the extreme degree of heterogeneity: every single neuron is different. Specifically, I write down an energy function and the corresponding dynamical equations for neurons' states assuming that each neuron has its own activation function and a kinetic time scale.  The network is assumed to be fully connected, so that every neuron is connected to every other neuron using a symmetric matrix of weights $W_{IJ}$, indices $I$ and $J$ enumerate different neurons in the network, see Fig.\ref{Fig:architecture}A.  The easiest way to mathematically formulate this problem is to define the network through a Lagrangian function $L(\{x_I\})$ that depends on the activities of all the neurons in the network. The activation function for each neuron is defined as a partial derivative of the Lagrangian  with respect to that neuron's activity.
\begin{equation}
    g_I = \frac{\partial L}{\partial x_I}
\end{equation}
From the biological perspective one can think about $g_I$ as an axonal output of the neuron $I$. In the simplest case, when the Lagrangian is additive for different neurons, this definition results in the activation that is a non-linear function of that neuron's activity. For non-additive Lagrangians this activation function can  depend on the activities of a group of neurons. For instance, it can contain contrastive (softmax) or divisive normalization, see for example models B and C in \cite{Krotov_Hopfield_2020}. The dynamical equations describing temporal evolution of a given neuron $I$ are given by 
\begin{equation}
    \tau_I \frac{dx_I}{dt} = \sum\limits_{J=1}^N W_{IJ} g_J - x_I \label{FC:equations}
\end{equation}
This equation belongs to the class of models called firing rate models in neuroscience. Each neuron $I$ collects the axonal outputs $g_J$ from all the neurons, weights them with the synaptic coefficients $W_{IJ}$ and produces its own time-dependent activity $x_I$. The temporal evolution has a time constant $\tau_I$, which in general can be different for every neuron. This network has a global energy function 
\begin{equation}
    E = \sum\limits_{I=1}^N x_I g_I - L - \frac{1}{2} \sum\limits_{I,J=1}^N g_I W_{IJ} g_J \label{FC:Energy}
\end{equation}
where the first two terms represent the Legendre transform of the Lagrangian function with respect to the neurons' activities. The temporal derivative of this energy function can be computed on the dynamical trajectories leading to (see Appendix A for details)
\begin{equation}
    \frac{dE}{dt} = - \sum\limits_{I,K=1}^N \frac{dx_I}{dt} M_{IK} \frac{dx_K}{dt}\leq 0, \ \ \ \ \text{where}\ \ \ \ M_{IK} = \tau_I  \frac{\partial^2 L }{\partial x_I \partial x_K}  \label{FC:Stability}
\end{equation}
The last inequality sign holds provided that the matrix $M_{IK}$ (or its symmetric part) is positive semi-definite. If, in addition to this, the energy function is bounded from below the non-linear dynamical equations are guaranteed to converge to a fixed point attractor state. The advantage of formulating this network in terms of the Lagrangian functions is that it makes it possible to easily experiment with different choices of the activation functions and different architectural arrangements of neurons. For all those flexible choices the conditions of convergence are determined by the properties of the matrix $M_{IJ}$ and the existence of the lower bound on the energy function.  

It was shown in \cite{Krotov_Hopfield_2016, Demircigil, Hochreiter, Krotov_Hopfield_2020} that the memory storage capacity of this system can be decoupled from the dimensionality of the input space, unlike the classical Hopfield Network.  For certain choices of the Lagrangians (activation functions) this memory storage capacity can be made very large, provided that there is a sufficiently large number of hidden neurons available.  

\section{Hierarchical Layered Networks}
\begin{figure}[t]
\begin{center}
\includegraphics[width = 0.9\linewidth]{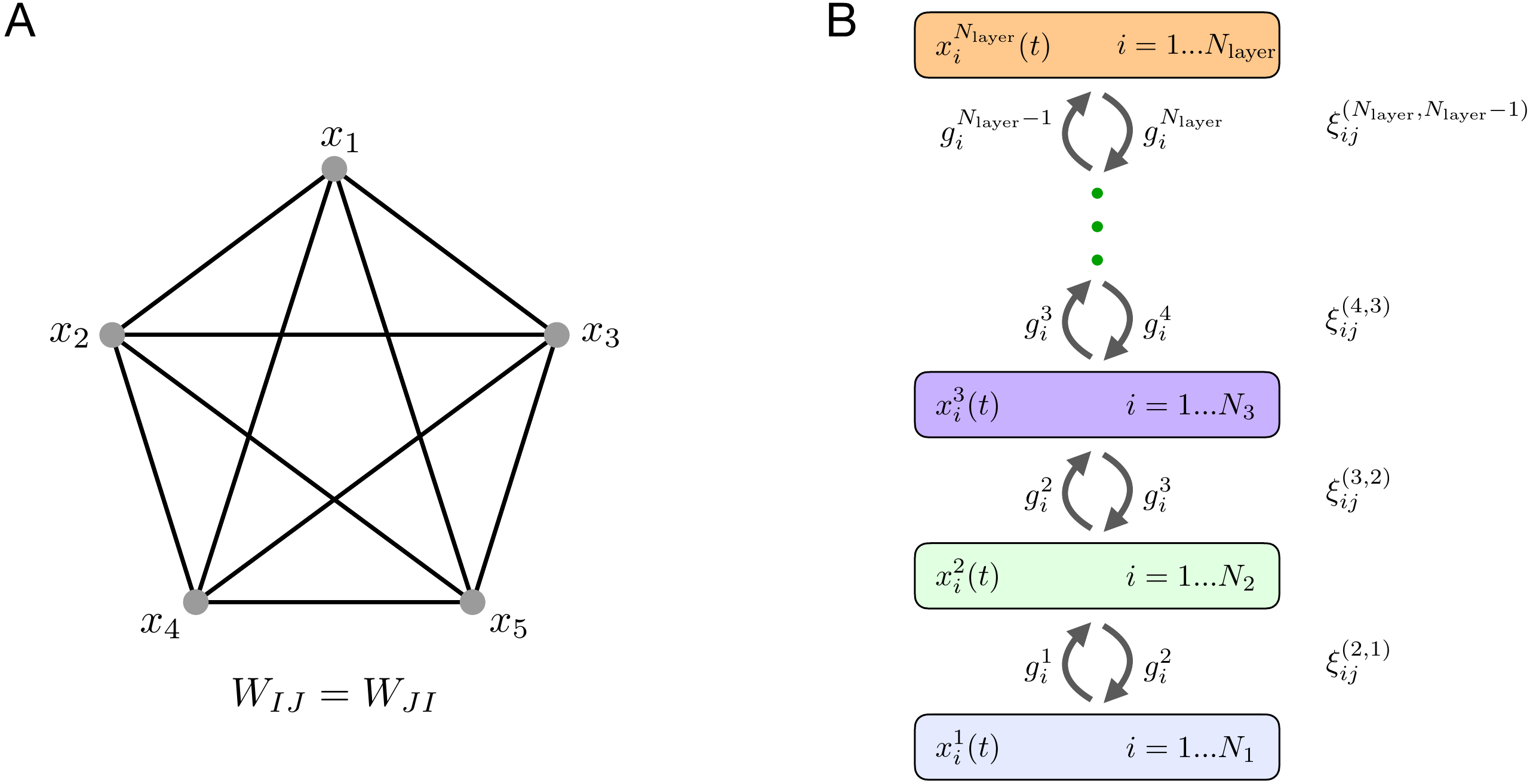}
\end{center}
\caption{(A) The connectivity diagram of the fully-connected network. The synaptic weights are described by the symmetric matrix $W_{IJ}$. (B) The connectivity diagram of the layered network. Each layer can have a different number of neurons, different activation function, and different time scales. The feedforward weights and feedback weights are equal.}\label{Fig:architecture}
\end{figure}
This section uses the general theory for fully connected networks presented in the previous section and organizes neurons in layers so that every neuron in a given layer has the same activation function and the same dynamic time scale. It is assumed that there are no horizontal connections between the neurons within the layer (lateral connections) and there are no skip-layer connections. Although these architectural features can be easily added, and the resulting dynamical equations and the energy function can be derived from (\ref{FC:equations}) and (\ref{FC:Energy}). A layered arrangement of neurons in networks with symmetric feedback weights (and an energy function) has been previously studied in \cite{Xie_Seung, EqProp}. Also, a layered network with feedback, but without an energy function has been studied in \cite{Fukushima}.

Consider an architecture consisting of $N_\text{layer}$ layers of recurrently connected neurons with the states described by continuous variables $x_i^{A}$ and the activation functions $g_i^{A}$, index $A$ enumerates the layers of the network, and index $i$ enumerates individual neurons in that layer. The activation functions in general can depend on the activities of all the neurons in the layer. The network diagram is shown in Fig.\ref{Fig:architecture}B. Every layer can have a different number of neurons $N_A$. These neurons are recurrently connected with the neurons in the preceding and the subsequent layers. The matrices of weights that connect neurons in layers $A$ and $B$ are denoted by $\xi^{(A,B)}_{ij}$ (the order of the upper indices for weights is the same as the order of the lower indices, in the example above this means that  index $i$ enumerates neurons in the layer $A$, and index $j$ enumerates neurons in the layer $B$). The feedforward weights and the feedback weights are equal, which is a consequence of the symmetry of the weights in equations (\ref{FC:equations}) and (\ref{FC:Energy}). The dynamical equations for the neurons' states can be written as \begin{equation}
    \tau_A \frac{dx_i^A}{dt} = \sum\limits_{j=1}^{N_{A-1}} \xi^{(A, A-1)}_{ij} g_j^{A-1} + \sum\limits_{j=1}^{N_{A+1}} \xi^{(A, A+1)}_{ij} g_j^{A+1} - x_i^A\label{Equations_of_motion}
\end{equation}
with boundary conditions 
\begin{equation}
    g_i^0 =0, \ \ \ \ \ \text{and}\ \ \ \ \ g_i^{N_\text{layer}+1}=0\label{Boundary_conditions}
\end{equation}
The main difference of these equations from the conventional feedforward networks is the presence of the second term, which is responsible for the feedback from higher layers. These top-down signals help neurons in lower layers to decide on their response to the presented stimuli. 
  
Following the general recipe from the previous section it is convenient to introduce a Lagrangian function $L^A(\{x^A_i\})$ for the $A$-th hidden layer, which depends on the activities of all the neurons in that layer. The activation functions in that layer can be defined as partial derivatives of the Lagrangian
\begin{equation}
    g_i^A = \frac{\partial L^A}{\partial x_i^A}\label{Def:activation functions}
\end{equation}
With these definitions the general energy (Lyapunov) function (\ref{FC:Energy}) for the network (\ref{Equations_of_motion}) reduces to 
\begin{equation}
    E = \sum\limits_{A=1}^{N_\text{layer}} \Big[ \sum\limits_{i=1}^{N_A} x_i^A g_i^A - L^{A}\Big] - \sum\limits_{A=1}^{N_\text{layer}-1} \sum\limits_{i=1}^{N_{A+1}} \sum\limits_{j=1}^{N_A} g_i^{A+1} \xi^{(A+1,A)}_{ij} g_j^A\label{Energy}
\end{equation}
If the Lagrangian functions, or equivalently the activation functions, are chosen in such a way that the Hessians for each layer are positive semi-definite and the overall energy is bounded from below, this system is guaranteed to converge to a fixed point attractor state. In Appendix A it is shown that the temporal derivative of this energy function is given by 
\begin{equation}
    \frac{dE}{dt} = -\sum\limits_{A=1}^{N_\text{layer}} \tau_A \sum\limits_{i,j=1}^{N_A} \frac{dx_j^A}{dt} \frac{\partial^2 L^{A}}{\partial x_j^{A} \partial x_i^{A}} \frac{dx_i^A}{dt} \leq 0 \label{Energy_decrease_layered}
\end{equation}
Thus, the network shown in Fig.\ref{Fig:architecture}B is indeed an attractor network with the global energy function. This network is described by a hierarchical set of synaptic weights $\xi^{(A+1,A)}_{ij}$ that can be learned for each specific problem. 

The network (\ref{Equations_of_motion}) has a structure of a recurrent network with a degree of biological plausibility. Each neuron in this network collects signals from the neurons in the layer above it, and below it, and produces its own state.
In the limit when the Lagrangian functions in each layer are additive for individual neurons 
\begin{equation}
    L^A = \sum\limits_{i=1}^{N_A} F(x_i^A)
\end{equation}
the axonal output of each neuron becomes a non-linear function of that neuron's activity
\begin{equation}
    g_i^A = F'(x_i^A)
\end{equation}
In this limit the network has only pair-wise interactions between the individual neurons. Thus, it has the same degree of biological plausibility as the single hidden layer Hopfield Network (modern or classical). This network also has an unbiological aspect, common to any Hopfield Network, which is that the feedforward weights and the feedback weights should be equal. If this constraint is violated it is impossible to derive an energy function, and, for this reason, the network can have more complicated dynamical trajectories than simple fixed-point attractors.

In the additive limit the Hessian in equation (\ref{Energy_decrease_layered}) becomes a diagonal matrix. The condition of positive semi-definiteness reduces to the requirement that the activation functions are monotonically growing. Thus, in this simple limit the HAM model reduces to the model studied in \cite{Xie_Seung}.  

\subsection{Hierarchical Time Scales, Adiabatic Limit}
Fixed points of the dynamical equations (\ref{Equations_of_motion}) and the energy function (\ref{Energy}) are independent of the time constants $\tau_A$. Thus, for a given set of weights the network will eventually arrive at the same fixed point (if the same initial conditions are given) regardless of the specific values of the time constants. At the same time, the dynamical trajectories (how the network arrives at that fixed point) can strongly depend on the choice of the time constants. 

Consider for example a self-supervised task, so that at the initial moment of time $t=0$ a noisy input is presented to the network, and the network has to remove the noise from the input as dynamics progresses. A possible way to solve this task is to initialize the input layer with the noisy input, and the hidden layers with zeros. Then let the network evolve with time until a fixed point is reached, and adjust the weights so that the activities of neurons in the input layer at the fixed point match the initial uncorrupted input.  In order for this network to operate as an associative memory, the information about the initial condition, noisy input presented at $t=0$ to the input layer, should have enough time to propagate through all the hidden layers. For this reason, it is convenient to assume that the dynamics of the hidden neurons in higher layers is always faster than the dynamics of the neurons in the lower layers, so that 
\begin{equation}
    \tau_{1}\gg\tau_2\gg ... \gg \tau_{N_\text{layer}}
\end{equation}
In physics jargon the input neurons evolve in time {\it adiabatically} (slowly) so that for every state of the input neurons hidden neurons have enough time to equilibrate. In most situations, see examples in the next section, it even makes sense to assume that the neurons in the top layer are instantaneously adjusting to the signals from the layer below, so that $\tau_{N_\text{layer}} =0$ (or very small). In this limit the expressions for the energy function can be simplified.

\section{Examples of Simple Architectures}
\begin{figure}[t]
\begin{center}
\includegraphics[width = 1.0\linewidth]{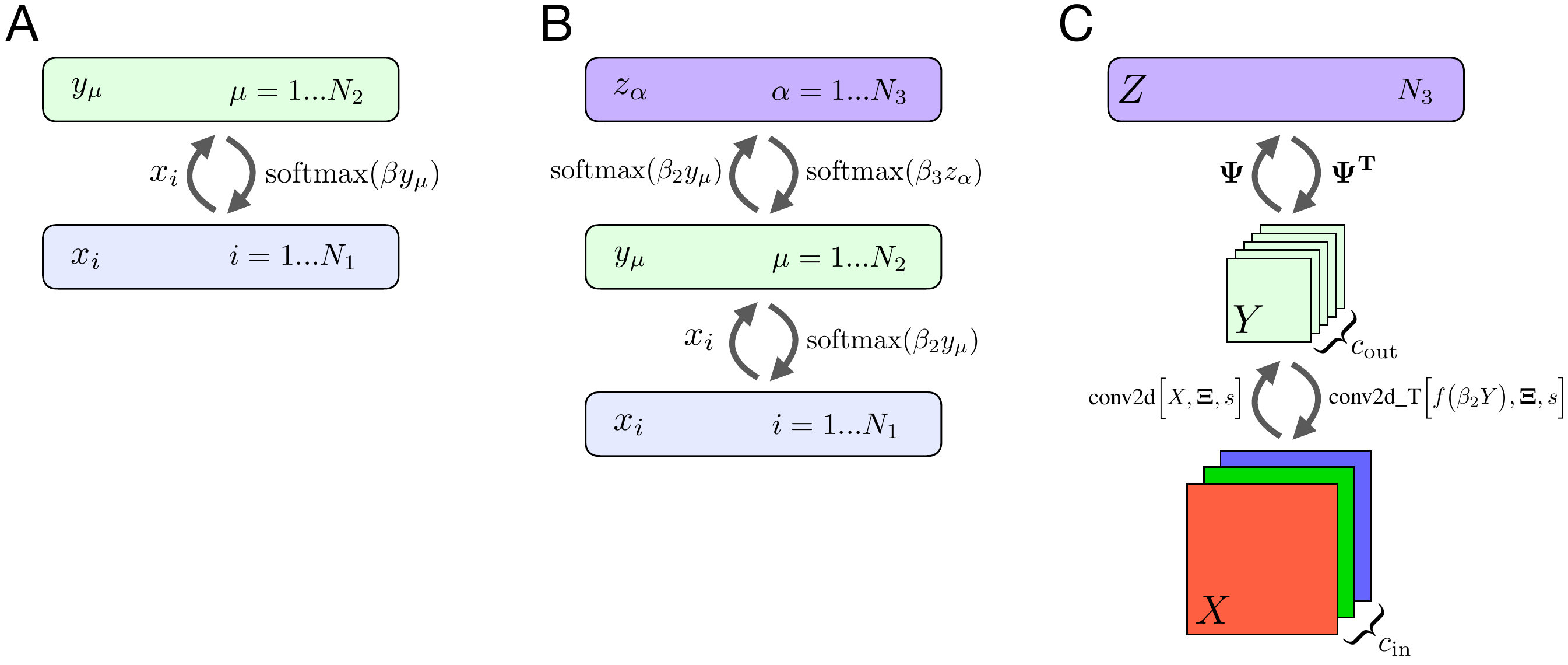}
\end{center}
\caption{(A) Modern Hopfield Network with one hidden layer. The output of the input layer is a linear function, the output of the hidden layer is a softmax with inverse temperature $\beta$. (B) HAM network with two hidden layers. The output of the input layer is a linear function, the output of the second layer is a softmax with inverse temperature $\beta_2$, the output of the third layer is a softmax with inverse temperature $\beta_3$.  (C) Convolutional HAM model with two hidden layers (convolutional and dense). }\label{Fig:simple_examples}
\end{figure}
This section describes three simple architectures, shown in Fig.\ref{Fig:simple_examples}, with gradually increasing complexity. For each network explicit dynamical equations and the corresponding energy function are derived. The first two architectures use dense layers, the last one uses a convolutional layer with shared weights. The goal of this section is to illustrate the application of the general formalism from the previous sections in these three simple examples.  

\subsection{HAM with One Hidden Layer}
First, as a warm-up exercise, consider a network with one hidden layer, see Fig.\ref{Fig:simple_examples}A. In this limit the HAM model reduces to the model previously studied in \cite{Hochreiter, Krotov_Hopfield_2020}. To simplify the notations compared to the general formulation (\ref{Equations_of_motion}) the activities of the neurons in the input layer are denoted by $x_i(t)$ and the activities of the hidden neurons by $y_\mu(t)$. Following the general recipe, the network is specified by the Lagrangian functions
\begin{equation}
    L^1 = \frac{1}{2} \sum\limits_{i=1}^{N_1} x^2_i, \ \ \ \ \ \ \ \ \ \ L^2 = \frac{1}{\beta} \log\Big(\sum\limits_{\mu=1}^{N_2} e^{\beta y_\mu}\Big)
\end{equation}
The activation functions are defined as partial derivatives of the Lagrangians through (\ref{Def:activation functions}). Thus, the axonal outputs of the first layer have simple linear activation functions, while the axonal outputs of the hidden neurons have a softmax activation $f\big(\beta y_\mu\big)$. This leads to the dynamical equations given by
\begin{equation}
\begin{cases}
    \begin{split}
        \tau_2\frac{dy_\mu}{dt} =& \sum\limits_{i=1}^{N_1}\Xi_{\mu i} x_i - y_\mu\\
        \tau_1\frac{dx_i}{dt}=& \sum\limits_{\mu=1}^{N_2}\Xi_{i\mu} f\big(\beta y_\mu\big) - x_i
    \end{split}
\end{cases}
\end{equation}
where the synaptic weights $\xi^{(2,1)}_{ij}$ between the input layer and the hidden layer are denoted by the matrix $\Xi_{\mu i }$, with the goal to simplify the notations in this case. The general formula for the energy function (\ref{Energy}) reduces to 
\begin{equation}
    E = \frac{1}{2} \sum\limits_{i=1}^{N_1} x^2_i + \sum\limits_{\mu=1}^{N_2} f\big(\beta y_\mu\big)y_\mu - \frac{1}{\beta} \log\Big(\sum\limits_{\mu=1}^{N_2} e^{\beta y_\mu}\Big) - \sum\limits_{\mu=1}^{N_2}\sum\limits_{i=1}^{N_1} f\big(\beta y_\mu\big) \Xi_{\mu i} x_i  \label{Energy:2layer}
\end{equation}
which monotonically decreases on the dynamical trajectories. If one further assumes that the network operates in the adiabatic limit $\tau_1\gg\tau_2$ and that $\tau_2=0$ (or very small), the equations for $y_\mu$ can be solved at the fixed point, which leads to the cancellation of the second and fourth terms in the energy function (\ref{Energy:2layer}). All these results have been extensively discussed in \cite{Krotov_Hopfield_2020}. 

\subsection{HAM with Two Dense Hidden Layers}
Now consider a network with two hidden layers, shown in Fig.\ref{Fig:simple_examples}B. All the layers are dense. The activities of neurons are denoted by $x_i$, $y_\mu$, and $z_\alpha$. The idea is to add one more hidden layer that communicates with the simple one-hidden-layer architecture through a softmax non-linearity. This leads to the following Lagrangians:
\begin{equation}
    L^1 = \frac{1}{2} \sum\limits_{i=1}^{N_1} x^2_i, \ \ \ \ \ \ \ \ \ \ L^2 = \frac{1}{\beta_2} \log\Big(\sum\limits_{\mu=1}^{N_2} e^{\beta_2 y_\mu}\Big), \ \ \ \ \ \ \ \ \ \ L^3 = \frac{1}{\beta_3} \log\Big(\sum\limits_{\alpha=1}^{N_3} e^{\beta_3 z_\alpha}\Big)
\end{equation}
Notice that the inverse temperature parameters $\beta$ for the axonal outputs of the first and second hidden layers can in general be different. They are denoted by $\beta_2$ and $\beta_3$ for the second and third layers. The equations (\ref{Equations_of_motion}) reduce to
\begin{equation}
\begin{cases}
    \begin{split}
        \tau_3 \frac{dz_\alpha}{dt} =& \sum\limits_{\mu=1}^{N_2}\Psi_{\alpha\mu} f\big(\beta_2 y_\mu\big) - z_\alpha \\
        \tau_2\frac{dy_\mu}{dt} =&\sum\limits_{\alpha=1}^{N_3} \Psi_{\alpha\mu} f\big(\beta_3 z_\alpha\big) + \sum\limits_{i=1}^{N_1}\Xi_{\mu i} x_i - y_\mu\\
        \tau_1\frac{dx_i}{dt}=& \sum\limits_{\mu=1}^{N_2}\Xi_{i\mu} f\big(\beta_2 y_\mu\big) - x_i 
    \end{split}
\end{cases}\label{Eqn:3layersDense}
\end{equation}
where the matrix of synaptic connections between the two hidden layers is denoted by $\Psi_{\alpha\mu}$, and the activation function $f(\cdot)$ is an element-wise softmax. Notice that the axonal outputs of the first hidden layer $f\big(\beta_2 y_\mu\big)$ provide both bottom-up signal to the second hidden layer (first term in the equation for $z_\alpha$), and top-down signal for the input layer (first term in the equation for $x_i$). The general expression for the energy function (\ref{Energy}) for this three-layer network reduces to
\begin{equation}
\begin{split}
    E = \frac{1}{2} \sum\limits_{i=1}^{N_1} x^2_i + \sum\limits_{\mu=1}^{N_2} f\big(\beta_2 y_\mu\big)y_\mu - \frac{1}{\beta_2} \log\Big(\sum\limits_{\mu=1}^{N_2} e^{\beta_2 y_\mu}\Big) + \sum\limits_{\alpha=1}^{N_3} p\big(\beta_3 z_\alpha\big)z_\alpha\\ - \frac{1}{\beta_3} \log\Big(\sum\limits_{\alpha=1}^{N_3} e^{\beta_3 z_\alpha}\Big) - \sum\limits_{\mu=1}^{N_2}\sum\limits_{i=1}^{N_1} f\big(\beta_2 y_\mu\big) \Xi_{\mu i} x_i - \sum\limits_{\alpha=1}^{N_3}\sum\limits_{\mu=1}^{N_2} p\big(\beta_3 z_\alpha\big) \Psi_{\alpha \mu} f\big(\beta_2 y_\mu\big) \label{Energy:3layersDense}
    \end{split}
\end{equation}
Furthermore, if we assume that the network operates in the adiabatic regime, and that the neurons in the top layer instantaneously adjust to the activities of the neurons in the lower two layers, so that $\tau_3=0$ (or very small), the energy function can be further simplified.  In this limit it reduces to 
\begin{equation}
    E = \frac{1}{2} \sum\limits_{i=1}^{N_1} x^2_i + \sum\limits_{\mu=1}^{N_2} f\big(\beta_2 y_\mu\big)y_\mu - \frac{1}{\beta_2} \log\Big(\sum\limits_{\mu=1}^{N_2} e^{\beta_2 y_\mu}\Big) - \frac{1}{\beta_3} \log\Big(\sum\limits_{\alpha=1}^{N_3} e^{\beta_3 z_\alpha}\Big) - \sum\limits_{\mu=1}^{N_2}\sum\limits_{i=1}^{N_1} f\big(\beta_2 y_\mu\big) \Xi_{\mu i} x_i 
\end{equation}
Since the Hessians of the Lagrangians are positive semi-definite, this quantity is guaranteed to decrease on the trajectory described by the non-linear dynamical equations. Since the weights of the network $W_{IJ}$ in equation (\ref{FC:equations}) are symmetric, the matrices $\mathbf{\Xi}$ and $\mathbf{\Psi}$ appear twice in the right hand side of equations (\ref{Eqn:3layersDense}): once in the forward path, and once in the feedback path.

\subsection{HAM with Two Hidden Layers and Local Connectivity}
The last example pertains to the network shown in Fig.\ref{Fig:simple_examples}C: it has two hidden layers, the first one is convolutional and the second one is dense. The intuitive motivation for this network is the following. Let's imagine that the inputs to this network are images, which we would like to store in the network as memories. The first hidden layer is then responsible for storing local patches of those images (the patches can be reused in multiple memories), while the second hidden layer is responsible for assembling those patches into images corresponding to the stored memories. 

This section is written in vector notations, as opposed to Einstein notations with explicit indices used in the previous sections. This is done for two reasons. First, the equations for the locally connected networks look simpler this way. Second, these notations make it easier to connect the proposed energy functions with the primary operations used in common machine learning frameworks. The coordinates of the pixels in the input layer $X$ are denoted by $x,y$ and the channel is denoted by index $\mu$, the coordinates in the feature map (first hidden layer $Y$) are denoted by $\Tilde{x}, \Tilde{y}$, and the channel is denoted by $\Tilde{\mu}$, parameters $w$ and $s$ denote the size of the convolutional window and the stride. 

As always, the starting point is the choice of the Lagrangian functions, which fully determine the dynamical equations and the energy function. They are given by 
\begin{equation}
    \begin{split}
        L^1 =& \frac{1}{2} \sum\limits_{x, y,\mu}  X^2_{x,y,\mu}\\
        L^2 =& \frac{1}{\beta_2} \sum\limits_{\Tilde{x},\Tilde{y}} \log\bigg( \sum\limits_{\Tilde{\mu}} e^{\beta_2 Y_{\Tilde{x},\Tilde{y}, \Tilde{\mu}}}\bigg) \\
        L^3 =& \frac{1}{\beta_3} \log\bigg( \sum\limits_{\alpha} e^{\beta_3 Z_{\alpha}}\bigg)
    \end{split}\label{Lagragians_conv}    
\end{equation}
and the corresponding activation functions can be computed using the definition (\ref{Def:activation functions}). While the equations for $L^1$ and $L^3$ are very similar to the dense case from the previous section, the equation for $L^2$ is different (notice the arrangement of summations). With this choice of the Lagrangian the softmax activation is taken at every position in the feature map plane $Y$ with respect to the channels. Thus at every position the channels (or filters corresponding to different channels) compete with each other.   With these notations the general dynamical equations (\ref{Equations_of_motion}) reduce to  
\begin{equation}
\begin{cases}
    \begin{split}
        \tau_3 \frac{dZ}{dt} = & \mathbf{\Psi}\  \text{flatten}\Big[f\big(\beta_2 Y\big)\Big] - Z \\
        \tau_2\frac{dY}{dt} = & \text{reshape}_{[\Tilde{L}, \Tilde{L}, c_\text{out}]}\Big[\mathbf{\Psi^T} p\big(\beta_3 Z\big)\Big] + \text{conv2d}\Big[X, \mathbf{\Xi}, s \Big] - Y\\
        \tau_1\frac{dX}{dt} = & \text{conv2d\textunderscore T} \Big[f\big(\beta_2 Y\big), \mathbf{\Xi}, s \Big] - X
    \end{split}\label{eqn_motion_conv}
\end{cases}
\end{equation}
where the activation function $f(\cdot)$ is a softmax along the channel dimension $\Tilde{\mu}$, $p(\cdot)$ is a softmax along the dimension of vector $Z$, and conv2d\textunderscore T is a 2d transposed convolution \cite{Dumoulin}. Assuming that the input images have the size $L\times L$ and the feature maps in the first convolutional layer have the size $\Tilde{L}\times\Tilde{L}$, the dimensions of all the tensors in the above equations are given by
\begin{equation}
    \begin{split}
        X:& \ [L, L, c_\text{in}]\\
        Y:& \ [\Tilde{L}, \Tilde{L}, c_\text{out}], \ \text{with}\ \ \Tilde{L} = \left \lfloor{\frac{L-w}{s}}\right \rfloor +1 \\
        Z:& \ [N_3]\\
        \mathbf{\Xi}:& \ [w,w,c_\text{in}, c_\text{out}]\\
        \mathbf{\Psi}:& \ [N_3, \Tilde{L}^2c_\text{out}]
    \end{split}
\end{equation}
The expression for the energy function for this model with convolutional layers can be derived by substituting definitions (\ref{Lagragians_conv}) into the general expression (\ref{Energy}). I will write it in the adiabatic limit, so that $\tau_1\gg\tau_2\gg\tau_3$, and assuming that $\tau_3=0$ so that the neurons in the top layer instantaneously react to the inputs from the second layer. In this case the equation for $Z_\alpha$ can be analytically solved, and the energy function (\ref{Energy}) reduces to
\begin{equation}
\begin{split}
    E =& \frac{1}{2} \sum\limits_{x,y,\mu} X^2_{x,y,\mu} + \sum\limits_{\Tilde{x},\Tilde{y}, \Tilde{\mu}} Y_{\Tilde{x},\Tilde{y}, \Tilde{\mu}} f\big(\beta_2 Y_{\Tilde{x},\Tilde{y}, \Tilde{\mu}}\big) -  \frac{1}{\beta_2} \sum\limits_{\Tilde{x},\Tilde{y}} \log\bigg( \sum\limits_{\Tilde{\mu}} e^{\beta_2 Y_{\Tilde{x},\Tilde{y}, \Tilde{\mu}}}\bigg)\\ &- \frac{1}{\beta_3} \log\bigg( \sum\limits_{\alpha} e^{\beta_3 Z_{\alpha}}\bigg) - \sum\limits_{\Tilde{x},\Tilde{y}, \Tilde{\mu}}  f\big(\beta_2 Y_{\Tilde{x},\Tilde{y}, \Tilde{\mu}}\big)\ \text{conv2d}\Big[X, \mathbf{\Xi}, s \Big]_{\Tilde{x},\Tilde{y}, \Tilde{\mu}}
\end{split}
\end{equation}
Note, that although matrix $\mathbf{\Psi}$ does not appear explicitly in this expression, the energy function implicitly depends on it through the dynamical variables. This energy function is guaranteed to decrease on the dynamical trajectory described by the equations (\ref{eqn_motion_conv}).  

Having derived the general formalism for networks with arbitrary large depth (\ref{Equations_of_motion}, \ref{Energy}), and the 3-layer example explicitly written in terms of convolutional arithmetic it is straightforward to stack an arbitrary large number of convolutional or dense layers on top of each other and write down the equations for the dynamics and the energy functions for those deeper models. 

{\bf Pooling Layer.} It is also possible to introduce an average pooling operation (but not max-pooling). The logic is similar to the convolutional layer, and the relationship between conv2d and conv2d\textunderscore T operations. In the forward path one applies the regular average pooling layer, which can be written as a linear operator acting on the flattened input. In the feedback path one applies the transpose of that operator used in the forward path. The resulting network will have an energy function. 

Convolutional continuous state Hopfield networks with neuron-wise activation functions have been previously  studied in \cite{Singer}. The expression for the energy function used in that work requires the inversion of the activation function, which is undefined for ``collective'' activation functions (like softmax) used in this paper. The energy function derived in the aforementioned work can be derived as a limiting case from equation (\ref{Energy}) in the convolutional setting and neuron-wise activation functions. The convolutional setting was also discussed in \cite{Ernoult}. The primitive function they develop, analog of the energy function (equation 18 in \cite{Ernoult}), however, is just a quadratic form. Dynamical equations defined as derivatives of that function are linear. Thus the aforementioned primitive function cannot describe non-linear dynamics of associative memory, which is the focus of the present work. 

\section{Discussion and Conclusions} The main contributions of this work are the following: 
\begin{enumerate}
    \item The general formulation of the fully-connected Modern Hopfield Network with arbitrary activation functions and arbitrary kinetic time constants is described (both neuron-wise non-linearities and activation functions involving groups of neurons are allowed). This network uses only local operations for the memory retrieval step, can operate in the regime of high memory load (far exceeding linear scaling with the number of input neurons), and has the same degree of biological plausibility as the classical Hopfield Network \cite{Hopfield_82,Hopfield_84}. This formulation relies on the recently proposed formalism of the Lagrangian functions \cite{Krotov_Hopfield_2020}. 
    \item This general formulation is used for constructing a hierarchical layered model of associative memory, that can have an arbitrary large number of hidden layers, arbitrary activation functions in every layer, and dense or local connectivity. Additionally, a possible hierarchical arrangement of kinetic time scales and the adiabatic limit is discussed.
    \item The general method for deriving the energy functions for all the considered models is proposed, and the conditions for the dynamics to arrive at the fixed point attractor state are formulated. These energy functions are non-linear expressions that describe the temporal evolution of neurons' states.  
\end{enumerate}

The biggest difference of the described models from the conventional feedfoward networks is the presence of top-down feedback pathway. From the computational perspective, this feedback helps lower layer neurons to decide on their response to local features, by providing a top-down signal that informs about more global features. For instance, if this network is applied to image processing tasks, it might be helpful for the feature detector, which recognizes an eye of an animal, to decide whether or not there is an eye in the newly presented image, if the top-down pathway communicates to it that there is a tail located somewhere else in that image. From the neurobiological perspective, the existence of feedback weights is a ubiquitous feature of biological circuits. In certain situations such feedback weights have a known computational function, see for example \cite{Feedback_function}. Introduction of feedback weights that are independent of the feedforward weights would break the symmetry of the network necessary for the existence of the energy function, and in many cases will result in computationally intractable behaviours with limit cycles, complicated attractors, or even chaotic behavior. For this reason, from the machine learning perspective, introduction of symmetric feedback weights might be the sweet spot intermediate step solution that brings the machine learning architectures one step closer to the biological reality (introduction of feedback), while still keeping the network in the mathematically tractable regime.        

Restricted to one hidden layer, the general models discussed in this work reduce to Dense Associative Memories previously studied in \cite{Krotov_Hopfield_2016, Demircigil, Hochreiter, Krotov_Hopfield_2020}. There is a mounting body of work \cite{Immune, Nguyen, Krotov_Hopfield_2018} suggesting that these networks can solve challenging machine learning problems in novel and beneficial ways. In those simpler models it is possible to find analytical expressions for the number of stored memories, assuming random ensembles of patterns. I am not aware of similar results for the models described in this work. This remains a challenging open problem. 

One obvious choice for training HAMs is the back propagation through time algorithm. There are other options. For instance a fixed-point training algorithm in the spirit of \cite{Pineda, Almeida} might be a more natural choice for certain problems. Other possible choices are: contrastive Hebbian learning \cite{Xie_Seung} or equilibrium propagation \cite{EqProp}. The use of higher order Runge-Kutta solvers, like in \cite{NeurODE, DEQ}, might also be beneficial. 

Last but not least, the HAMs models are more advanced (compared to previously published work), but still simple examples of general recurrent networks with the energy function. Using the formalism described in this work it is possible to derive many more novel architectures.  For instance, it is straightforward to introduce lateral connections within each layer, or skip connections. A recent work \cite{Tang} pointed out the relationship between mixer architectures \cite{Mixer1,Mixer2} and Model C of \cite{Krotov_Hopfield_2020}, which is a limiting case of HAMs. It is also possible to design networks with gated units. In all these applications the general formalism with the Lagrangian functions, see section \ref{FC:Section} and \cite{Krotov_Hopfield_2020}, should be helpful for designing more general neural architectures. 

\section*{Acknowledgements} I want to thank J.Hopfield and X.You for helpful discussions.

\section*{Appendix}
In this Appendix it is shown that the energy function (\ref{FC:Energy}) is a Lyapunov function for the fully connected Modern Hopfield Network (\ref{FC:equations}). And the same statement for the Hierarchical Associative Memory Network defined by the equations (\ref{Energy}) and (\ref{Equations_of_motion}). 

For the fully-connected network the time derivative of the energy (\ref{FC:Energy}) can be calculated as (the brackets are used to group the derivatives of the three terms in the energy function)
\begin{equation}
    \begin{split}
        \frac{dE}{dt} =& \bigg(\sum\limits_I \frac{dx_I}{dt}\frac{\partial L}{\partial x_I}  + \sum\limits_{I,K} x_I \frac{\partial^2 L}{\partial x_I \partial x_K} \frac{dx_K}{dt}\bigg) - \bigg(\sum\limits_I \frac{\partial L}{\partial x_I} \frac{dx_I}{dt}\bigg) \\& - \bigg(\sum\limits_{I,J,K} \frac{dx_K}{dt} \frac{\partial^2L }{\partial x_K \partial x_I} W_{IJ} g_J\bigg) =
        -\sum_{K,I} \frac{dx_K}{dt} \frac{\partial^2 L}{\partial x_K \partial x_I} \Big[ \sum\limits_J W_{IJ} g_J - x_I\Big] = \\ & - \sum_{K,I} \frac{dx_K}{dt} \frac{\partial^2 L}{\partial x_K \partial x_I} \tau_I \frac{dx_I}{dt} \leq 0
    \end{split}
\end{equation}
The first term in the first bracket cancels the second bracket, which leads to equation (\ref{FC:Stability}) from the main text. 

For the layered network (\ref{Energy}) and  (\ref{Equations_of_motion}) an analogous computation of the temporal derivative of the energy function gives
\begin{equation}
\begin{split}
    \frac{dE}{dt} &= \sum\limits_{A=1}^{N_\text{layer}} \sum\limits_{i,j=1}^{N_A} x_i^A \frac{\partial^2 L^A }{\partial x_i^A \partial x_j^A} \frac{dx_j^A}{dt}  - \sum\limits_{A=1}^{N_\text{layer}-1} \sum\limits_{i,j=1}^{N_{A+1}}\sum\limits_{k=1}^{N_A} \frac{dx_j^{A+1}}{dt}\frac{\partial^2 L^{A+1}}{\partial x_j^{A+1} \partial x_i^{A+1}}\xi^{(A+1,A)}_{ik} g_k^A \\ &- \sum\limits_{A=1}^{N_\text{layer}-1} \sum\limits_{k=1}^{N_{A+1}}\sum\limits_{i,j=1}^{N_A} g_k^{A+1}\xi^{(A+1,A)}_{ki} \frac{\partial^2 L^{A}}{\partial x_i^{A} \partial x_j^{A}} \frac{dx_j^A}{dt} =  \sum\limits_{i,j=1}^{N_1} \frac{dx_j^1}{dt}\frac{\partial^2 L^1}{\partial x_j^1 \partial x_i^1} \Big[ x_i^1- \sum\limits_{k=1}^{N_2} \xi^{(1,2)}_{ik} g_k^2 \Big]\\& + 
    \sum\limits_{i,j=1}^{N_\text{layer}} \frac{dx_j^{N_\text{layer}}}{dt}\frac{\partial^2 L^{N_\text{layer}}}{\partial x_j^{N_\text{layer}} \partial x_i^{N_\text{layer}}} \Big[ x_i^{N_\text{layer}}- \sum\limits_{k=1}^{N_\text{layer}-1} \xi^{(N_\text{layer},N_\text{layer}-1)}_{ik} g_k^{N_\text{layer}-1} \Big]\\& + \sum\limits_{A=2}^{N_\text{layer}-1}
    \sum\limits_{i,j=1}^{N_A} \frac{dx_j^{A}}{dt}\frac{\partial^2 L^{A}}{\partial x_j^{A} \partial x_i^{A}} \Big[ x_i^{A}- \sum\limits_{k=1}^{N_{A-1}} \xi^{(A,A-1)}_{ik} g_k^{A-1} - \sum\limits_{k=1}^{N_{A+1}} \xi^{(A,A+1)}_{ik} g_k^{A+1} \Big] = \\& -\sum\limits_{A=1}^{N_\text{layer}} \tau_A \sum\limits_{i,j=1}^{N_A} \frac{dx_j^A}{dt} \frac{\partial^2 L^{A}}{\partial x_j^{A} \partial x_i^{A}} \frac{dx_i^A}{dt} \leq 0
\end{split}
\end{equation}
After the second equality sign results of the differentiation are split into the first and last layer terms, which are different from the intermediate layers due to boundary conditions (\ref{Boundary_conditions}), and the intermediate layers. The expressions in the square brackets correspond to the right hand sides of the dynamical equations (\ref{Equations_of_motion}). In the last equality sign the right hand sides of those equations are used to replace expressions in the square brackets by the corresponding time derivatives of the neuron's activities. This completes the proof that the energy function decreases on the dynamical trajectory described by equations (\ref{Equations_of_motion}) for arbitrary time constants $\tau_A$ and Lagrangian functions (equivalently activation functions) provided that the Hessians of these Lagrangian functions are positive semi-definite. In order for the network to converge to a fixed point the Lagrangian functions should also be chosen in such a way that the energy function is bounded from below.


\end{document}